\documentclass[sigconf]{acmart}

\AtBeginDocument{%
  \providecommand\BibTeX{{%
    \normalfont B\kern-0.5em{\scshape i\kern-0.25em b}\kern-0.8em\TeX}}}

\setcopyright{acmcopyright}
\copyrightyear{2020}
\acmYear{2020}
\acmDOI{10.1145/1122445.1122456}

\acmConference[Petra '21]{Petra '21: The Pervasive Technologies Related to Assistive Environments}{June 29-- July 02, 2021}{Corfu, Greece}
\acmBooktitle{Petra '21: The Pervasive Technologies Related to Assistive Environments,
  June 29-- July 02, 2021, Corfu, Greece}
\acmPrice{15.00}
\acmISBN{978-1-4503-XXXX-X/18/06}

\begin{document}

\title{Self-Supervised Human Activity Recognition by Augmenting Generative Adversarial Networks}


\author{Mohammad Zaki Zadeh}
\affiliation{%
  \institution{University of Texas at Arlington, USA}
  \city{Arlington}
  \country{USA}}
\email{mohammad.zakizadehgharie@mavs.uta.edu}

\author{Ashwin Ramesh Babu}
\affiliation{%
  \institution{University of Texas at Arlington}
  \city{Arlington}
  \country{USA}}
\email{ashwin.rameshbabu@mavs.uta.edu}

\author{Ashish Jaiswal}
\affiliation{%
  \institution{University of Texas at Arlington}
  \city{Arlington}
  \country{USA}}
\email{ashish.jaiswal@mavs.uta.edu}

\author{Fillia Makedon}
\affiliation{
  \institution{University of Texas at Arlington}
  \city{Arlington}
  \country{USA}}
\email{makedon@uta.edu}

\renewcommand{\shortauthors}{Zaki Zadeh, et al.}

\begin{abstract}
This article proposes a novel approach for augmenting generative adversarial network (GAN) with a self-supervised task in order to improve its ability for encoding video representations that are useful in downstream tasks such as human activity recognition. In the proposed method, input video frames are randomly transformed by different spatial transformations, such as rotation, translation and shearing or temporal transformations such as shuffling temporal order of frames. Then discriminator is encouraged to predict the applied transformation by introducing an auxiliary loss. Subsequently, results prove superiority of the proposed method over baseline methods for providing a useful representation of videos used in human activity recognition performed on datasets such as KTH, UCF101 and Ball-Drop. Ball-Drop dataset is a specifically designed dataset for measuring executive functions in children through physically and cognitively demanding tasks. Using features from proposed method instead of baseline methods caused the top-1 classification accuracy to increase by more then $4\%$. Moreover, ablation study was performed to investigate the contribution of different transformations on downstream task.
\end{abstract}



\maketitle

\section{Introduction}

\begin{figure*}[ht]
\begin{center}
    \includegraphics[width=1\linewidth]{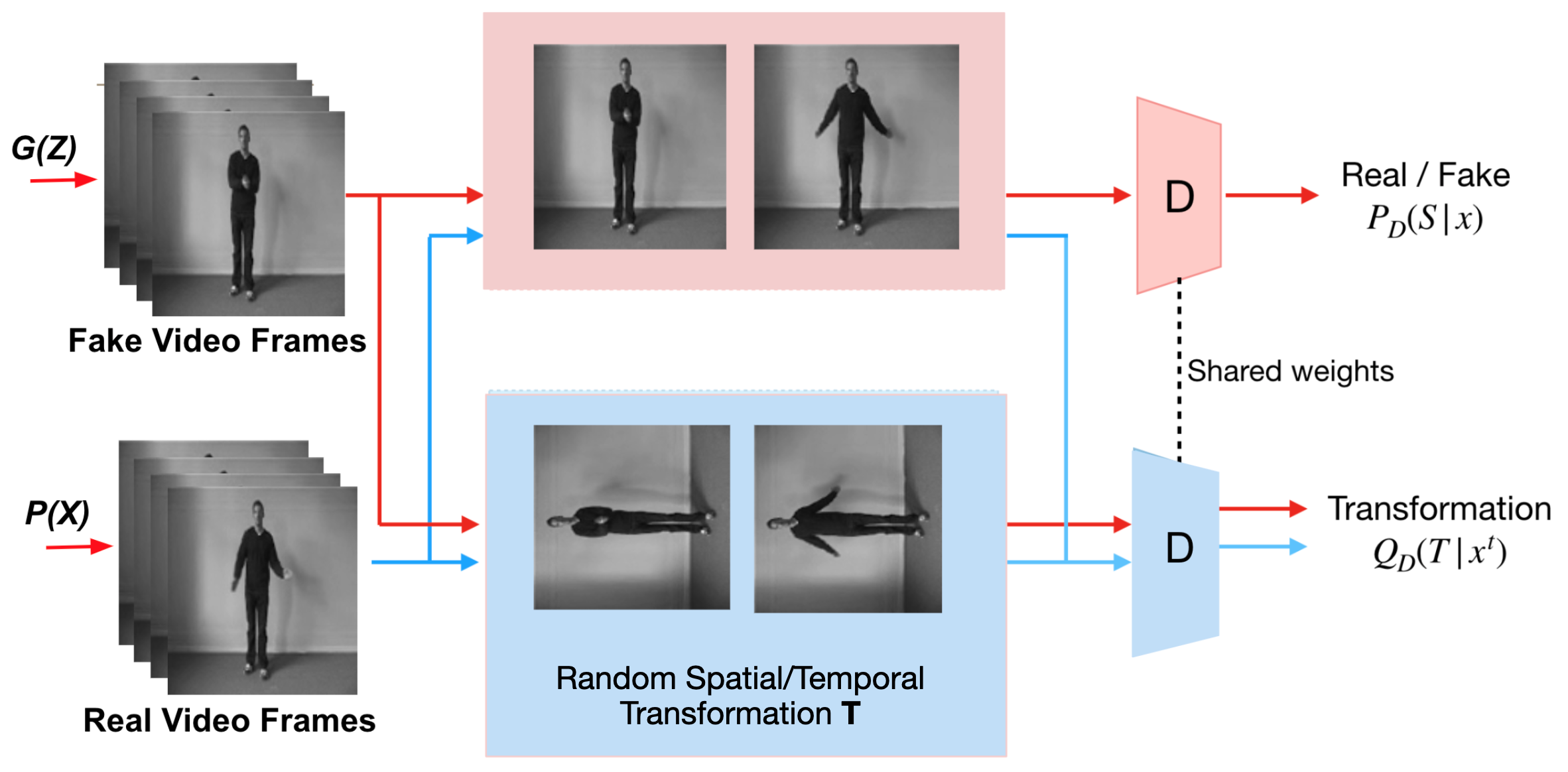}
\end{center}
  \caption{Proposed method architecture. Diagram is inspired by \cite{chen18}}
\label{fig:arc}
\end{figure*}

Recent advances in Deep Learning \cite{lecun15} and the challenge of gathering huge amounts of labeled data have encouraged new research in unsupervised or self-supervised learning. In particular, Computer Vision tasks could greatly benefit from successful models that learned abstract low-dimensional features of images and videos without any supervision, because unlabeled images and video sequences can be gathered automatically without human intervention ~\cite{ahsa18, kong18, cast18}.

As a result, a lot of research effort has been focused on methods that can adapt to new conditions without expensive human supervision. The main focus of this paper is on self-supervised visual representation learning, which is a subclass of unsupervised learning. Self-supervised learning techniques have produced state of the art low-dimensional representations on most computer vision benchmarks ~\cite{doer15, chen18, noro16, trin19, chen20}.

In self-supervised learning framework, only unlabeled data is needed in order to formulate a learning task, such as predicting context ~\cite{doer15} or image rotation ~\cite{chen18} for which a target objective can be computed without supervision. These methods usually incorporate Convolutional Neural Networks  (CNN) \cite{alex12} which after training, their intermediate layers encode high-level semantic visual representations. The obtained representations can be used for solving downstream tasks of interest, such as object detection or human activity recognition. Moreover Self-supervised learning can be employed in finding internal representations of the environment, which is useful in model-based reinforcement learning settings ~\cite{kais19}. 

While most of the research in application of self-supervised learning in computer vision is concentrated on still images, the focus of this paper is human activity recognition in videos. This work is motivated by the real-world ATEC (Activate Test of Embodied Cognition) system \cite{dill19,babu18,icmi20}, which assesses executive function in children through physically and cognitively demanding tasks. The system requires manually labeling hundred of hours of videos by experts. Therefore, this work exploits self-supervised learning techniques to extract low-dimensional representations of the videos. 

The method proposed in this work (Figure \ref{fig:arc}) is inspired by \cite{chen18} that augments Generative Adversarial Networks (GAN) ~\cite{good16, rad15} with self-supervised rotation loss in order to improve the representation capability of discriminator network. However, this work has significant differences. First, the purpose of this work is to find a low-dimensional representation of videos rather than still images. Second, In \cite{chen18} an auxiliary loss is added to the discriminator network to detect angles of random rotation applied on still images.
But in this work, the discriminator classifies among three different spatial transformations, such as rotation, translation or shearing and a temporal transformation that shuffles the temporal order of frames. All aforementioned transformations are randomly applied on video frames. Moreover, a thorough ablation study is performed which investigates the effect of each different transformation. 

The result prove that in general the introduction of self-supervised transformation loss, improves the quality of representation provided by discriminator network. It also shows that compare to baseline self-supervised GAN \cite{chen18}, inclusion of additional spatial transformations and temporal shuffling improves the downstream classification accuracy specially in Ball-Drop dataset. The rest of the paper is organized as follows: In section 2 a brief review of some of the recent self-supervised methods used in computer vision is provided. In section 3 the mathematical basis of self-supervised methods employed in this article is discussed. Then in section 4, results of proposed methods along with datasets and criteria used is presented. Finally, the last section includes, the conclusion and directions for future research are discussed.

\begin{figure*}[ht]
\begin{center}
    \includegraphics[width=0.8\linewidth]{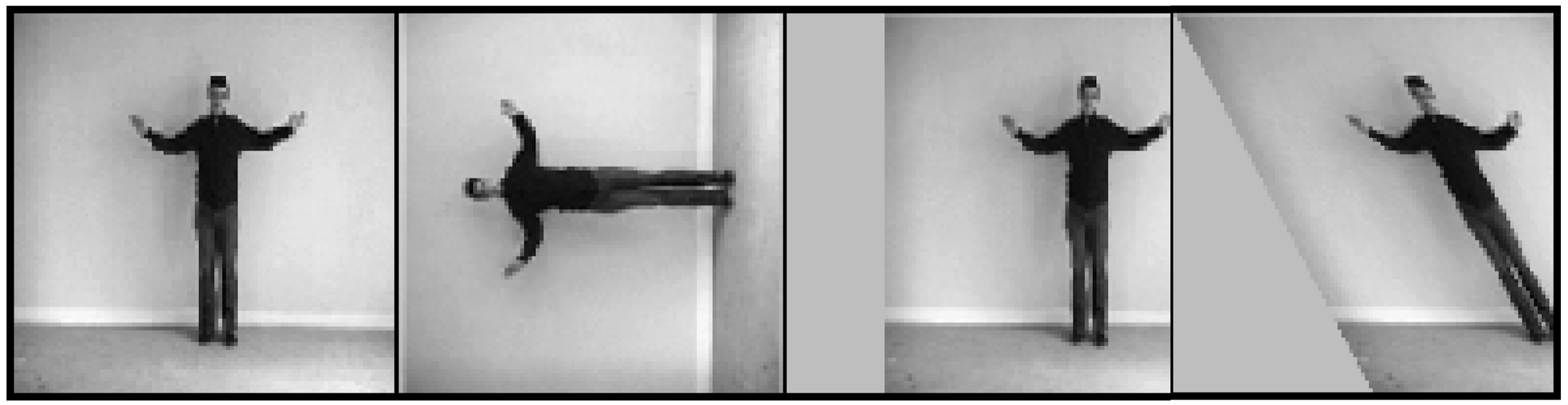}
\end{center}
  \caption{Examples of spatial transformation used. From left to right: Original Image, rotation, translation, shear.}
\label{fig:trans}
\end{figure*}

\begin{figure}[ht]
\begin{center}
    \includegraphics[width=0.9\linewidth]{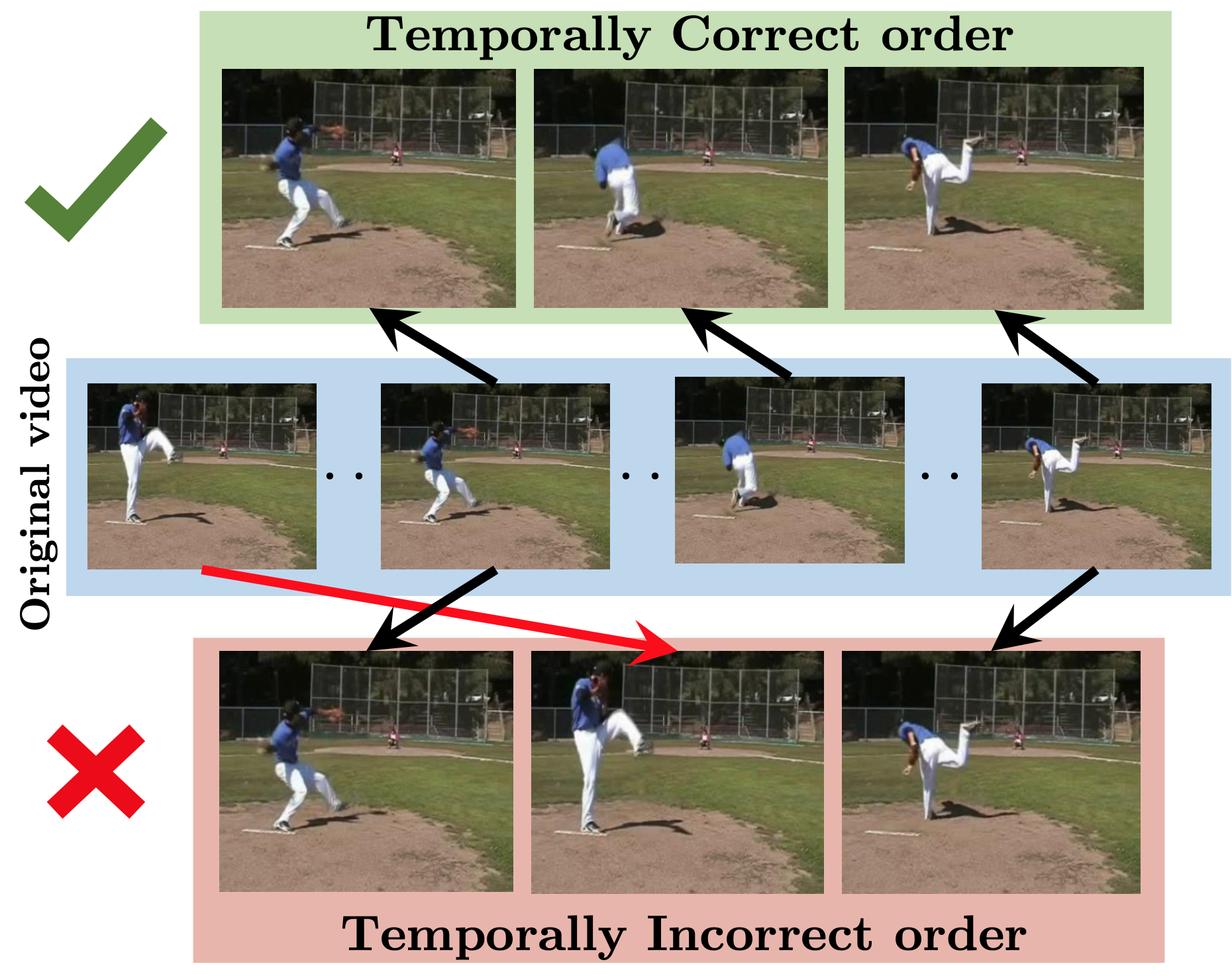}
\end{center}
  \caption{In temporal transformation used in proposed method, classifier tries to find if temporal order of videos frames are shuffled or not (Adopted from \cite{misr16}).
  }
\label{fig:temp}
\end{figure}

\section{Related Works}

In this section, some of the state of the art research works that employed self-supervised learning framework in computer vision are briefly summarized. Authors in \cite{doer15} the use of spatial context as supervisory signal was explored for learning image representation. For training, the authors selected two random pairs of patches from each image and tried to predict the relative position of the second patch in respect to the first one. Other efforts  used colorizing gray-scale images \cite{zhan16} and reconstructing missing parts of an image (Image Inpainting) \cite{path16} as self-supervised task for learning features. Researchers in \cite{noro17} counting the number of visual primitives in images is considered as a supervision signal. This signal is acquired without any manual annotation by using equivariance relations. Authors in \cite{noro16} divided an image into 9 tiles and shuffled their position via a randomly chosen permutation from a predefined permutation set and then predicted the index of the chosen permutation. All of these patches are sent through the same network, then their representations are concatenated and passed through fully-connected multi-layer perceptrons for prediction.

Another useful supervision signal is rotation. In \cite{gida18} the authors randomly rotated an input image and trained a deep convolutional neural networks to predict the rotation angle. In a similar fashion, authors in \cite{chen18} augmented a generative adversarial network with a rotation loss that encourages discriminator to classify which rotation was performed on input image. Contrastive Learning \cite{chen20,mdpi20} is another interesting idea that has achieved state of the art results. In contrast to generative models that generate computationally expensive pixel-level images, contrastive learning methods perform self-supervised task in latent space. For example in \cite{trin19} given masked-out patches in an input image, Contrastive Predictive Coding loss is used to learn to select the correct patch, among other distractor patches sampled from the same image to fill in the masked location. The authors employ a network of convolutional blocks to process patches followed by an attention pooling network to encode the content of unmasked patches before predicting masked ones. 
Furthermore self-supervised learning has been very successful in finding representation of videos. One of the most popular self-supervised task for video is prediction of future frames given past frames \cite{mat15,aign18}. In \cite{path16v, han19} it was proven that predicting future frames or motion of objects can provide a compact representation of video that can be exploited in downstream tasks. Another interesting task is utilizing temporal order of frames by either detecting whether temporal order of frames is valid \cite{misr16} or sorting shuffled frames \cite{Lee17}.

The method proposed in this work is also focused on providing low-dimensional representation of videos. It tries to predict which random transformation, spatial or temporal has been applied on input video frames.

\begin{figure*}[ht]
\begin{center}
    \includegraphics[width=0.8\linewidth]{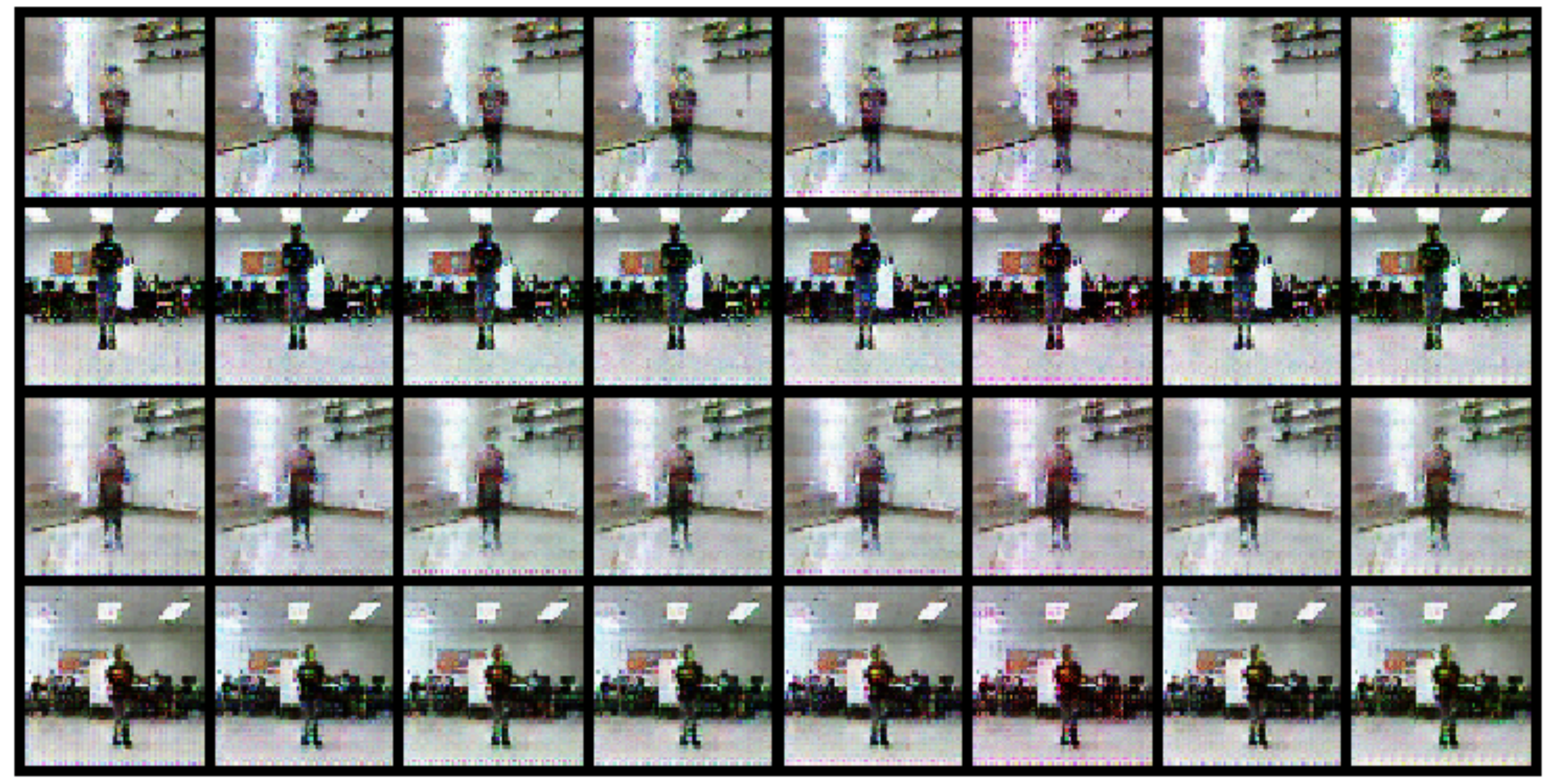}
\end{center}
  \caption{Samples of generated images from custom Ball-Drop dataset.}
\label{fig:ballDrop}
\end{figure*}

\section{Methodology}

In this section, first GAN is introduced which is the basis of methods used in this work. Subsequently the proposed self-supervised GAN is described in detail and how video representation (features) are extracted from it. These features are fed into a simple 2 layer multi-layer perceptrons (MLP) network for downstream classification tasks such as human activity recognition.

\subsection{GAN}
GAN ~\cite{good16, rad15} is a framework for producing a model distribution that mimics a given target distribution, and it consists of a generator $G(z;\theta_g)$ that produces the model distribution and a discriminator $D(x; \theta_d)$ that distinguishes the model distribution from the target. Training data is denoted by $x$ and input noise is $z$ with probability distribution of $P_z(z)$.

In practice both generator and discriminator are implemented by  differentiable CNNs with parameters: $\theta_g$ and $\theta_d$. $D$ is trained to maximize the probability of assigning the correct label to both training examples and samples from $G$. At the same time $G$ is trained to minimize $log(1-D(G(z)))$. In other words, $D$ and $G$ play the following two-player minimax game with value function $V(D, G)$ :

\begin{equation}\label{eq:eq1}
\begin{split}
\min_G \max_D V(D,G) =
\mathbb{E}_{x \sim P_{data}(x)}[log D(x)] \\ +  \mathbb{E}_{z \sim P_z(z)} [log (1-D(G(z)))].
\end{split}
\end{equation}

But using GAN in practice is challenging because of instability in training, mode collapse, etc. However in recent years variety of novel techniques such as gradient penalty \cite{gul17} or spectral normalization \cite{miya18} have been proposed to solve some of the challenges.

\subsection{Self-supervised Learning}
One of the main problems with GANs that limits their ability for providing good representation is discriminator forgetting \cite{chen18}. Because in practice as parameters of generator $G$ varies so does the distribution $P_G$ which causes learning process of discriminator to be non-stationary. In other words, the discriminator is not encouraged to keep a useful data representation as long as the current representation is useful to discriminate between the classes.

So in order to alleviate this problem, the discriminator network is augmented with a self-supervised task like predicting rotation angle \cite{gida18} or counting objects in image \cite{noro17} to motivate GAN to learn a useful compact representations. The method proposed in this work is based on spatial and temporal transformation of video frames. In this method (Figure \ref{fig:arc}), one transformation is randomly picked and applied on frames of input video. Then the self-supervised task is predicting the transformation used on video frames. As a result the loss function of both generator and discriminator are modified as follows:

\begin{equation}\label{eq:eq2}
\begin{split}
L_G = -V(D,G) - \alpha \mathbb{E}_{x \sim P_G} \mathbb{E}_{t \sim T} [log Q_D(T=t|x^t)] \\
L_D = V(D,G) - \mathbb{E}_{x \sim P_{data}} \mathbb{E}_{t \sim T} [log Q_D(T=t|x^t)]
\end{split}
\end{equation}

where $V(D,G)$ is the value function from Equation 1 and  $t \in T$
is a transformation selected from a set of possible spatial and temporal transformations. $x^t$ is input $x$ transformed by transformation $t$, $Q_D(T|x^t)$ is discriminator distribution over possible transformations and $\alpha$ is self-supervised loss weight. For this method three different spatial affine transformations such as rotation, translation and shearing along with a temporal transformation, in which temporal order of video frames are shuffled, are chosen. Example of spatial and temporal transformations are depicted in Figure \ref{fig:trans} and \ref{fig:temp} respectively.

For rotation only four classes were considered corresponding to rotation angles of $0^\circ$, $90^\circ$, $180^\circ$ and $270^\circ$. Respectively, three classes for translation (vertical, horizontal and both), three for shearing (vertical, horizontal and both) and one class for temporal transformation (shuffled or not) were chosen. So in total eleven different transformation classes were selected. 

As explained by \cite{chen18}, generator and discriminator are collaborative with respect to predicting the transformation task. Because for detecting the transformations, the  discriminator is trained only on the true data thus the generator is motivated to generate images that are easy for discriminator to detect. As illustrated in Figure \ref{fig:arc} the discriminator has two heads, which the former like normal GANs predicts whether non-transformed video frames are real or fake. The latter head on the other hand predicts the transformation class of transformed inputs.

After training is completed, output of the last layer before the heads is extracted as a compact representation of the input video. Then a simple 2 layer feed forward MLP is trained on extracted video representations for human activity recognition.

\begin{table*}
\begin{center}
\begin{tabular}{|l|c|c|c|}
\hline
Method & KTH & UCF101 & Ball-Drop \\
\hline\hline
GAN & $71.46\pm2.5$ & $64.68\pm0.4$ & $77.93\pm2.7$ \\
GAN+Rotation & $74.47\pm2.5$ & $66.86\pm0.6$ & $80.47\pm2.5$\\
GAN+Spatial & $76.41\pm2.0$ & $66.95\pm1.6$ & $81.99\pm4.5$\\
GAN+Temporal & $76.09\pm3.2$ & $70.88\pm0.7$ & $80.69\pm3.7$\\
GAN+SpatioTemporal & $77.13\pm3.6$ & $69.17\pm1.8$ & $84.53\pm3.0$\\
\hline
\end{tabular}
\end{center}
\caption{Top 1 classification accuracy of using features extracted from different methods.}
\label{results}
\end{table*}

\begin{table}
\begin{center}
\begin{tabular}{|l|c|c|c|}
\hline
Method &  Ball-Drop \\
\hline\hline
GAN & $77.93\pm2.7$  \\
GAN+Rotate & $80.47\pm2.5$  \\
GAN+Translate & $80.04\pm3.3$  \\
GAN+Shear & $79.52\pm3.3$  \\
GAN+Rotate+Translate & $81.32\pm5.1$  \\
GAN+Translate+Shear & $80.33\pm3.1$  \\
GAN+Rotate+Shear & $81.01\pm4.6$  \\
GAN+Rotate+Translate+Shear & $81.99\pm4.5$  \\
\hline
\end{tabular}
\end{center}
\caption{Investigating the effects of combination of different spatial transformations on Top 1 classification accuracy.}
\label{ablation}
\end{table}

\section{Result and Discussion}

In this section, first details of datasets used in this experiment are discussed. Then it is followed by discussion of neural network models used and how they are trained. Finally results of both baseline and proposed method are presented. It should be noted that in this article the focus is on providing compact representation of videos that can be exploited for activity recognition, thus evaluating fidelity of generated image frames is outside scope of this paper and as a result criteria such as Frechet Inception Distance (FID) are not used.     

\subsection{Datasets}
In this article in order to evaluate the performance of the proposed method for providing video representation useful for activity recognition three different video datasets were used. First two are publicly available video datasets like KTH \cite{kth} and UCF101 \cite{ucf101} containing short video clips of humans doing various activities. The third dataset which for simplicity is  called Ball-Drop (Ball-Drop-to-the-Beat) is based on one of tasks designed for ATEC system (Activate Test of Embodied Cognition) to assess both audio and visual cue processing of children while performing upper-body movements. The ATEC is an assessment test designed to measure executive functions in children through physically and cognitively demanding tasks \cite{dill19,babu18,icmi20}. 

For this task the child is required to pass a ball from one hand to another, following audio and visual cues. Based on the instructions, the child has to pass the ball for Green-Light, keep the ball still for Red-Light, and move the ball up and down with the same hand for Yellow-Light. There are 10 different tasks based on how the light colors are presented audibly or visually. There are total of 30 subjects present in this experiment, each recording 2 versions of all 10 different tasks. Each task consist of either 8 or 16 short segments that each one should be classified into 3 different classes of green-light, red-light and yellow-light. One of the main reasons that motivated authors of this article to pursue self-supervised learning is that manually annotating this dataset proved to be cumbersome and error prone. 

All of datasets used in this article were divided into 3 different sets. First 80\% of each dataset was considered as unlabeled and used solely for training self-supervised GANs. After training the remaining 20\% (labeled) were fed into trained discriminator network to extract video representations (features). Then again for activity recognition the features were divided into train and test set with ratio of 4 to 1.

\subsection{Models}
In self-supervised GANs for both generator and discriminator a 6 layer convolutional neural net (CNN) was used. Since the input is video, in discriminator the first 2 layer and for generator the last 2 layers employ 3D convolutional nets \cite{tran18}. As discussed by \cite{luci17,kura18}  performance of GANs depends on many different hyper-parameters and there is no set of hyper-parameters that guarantee superior performance on all datasets and finding one require massive computational budget. Due to our limited computational budget, very deep complex networks such as densenet and resnet101 were avoided \cite{hara18} and a small grid search was performed for tuning the hyper-parameters. 

All the models, including baseline GAN and proposed self-supervised GAN were trained for 100 epochs using PyTorch framework \cite{pytorch}  with ADAM \cite{adam} as optimizer with following parameters, which are selected empirically; 

generator learning rate: $0.0001$, generator learning rate: $0.0004$, beta1: $0.5$, beta2: $0.999$. Spectral Normalization was used in all methods to stabilize the training process. And for self-supervised GAN parameter of $\alpha$ in equation \ref{eq:eq2} was chosen as 0.25. Finally for doing classification on extracted features a 2 layer MLP were trained with ADAM optimizer with similar hyper-parameters. 

\subsection{Experimental Results}
In Figure \ref{fig:ballDrop} examples of generated images by proposed method are depicted. As stated at the start of this section, the quality of the generated images is not the focus of this paper. The real pictures of the children cannot be shown due to the privacy protection of the participants in Ball-Drop task. However, their generated images can be portrayed since faces of children are anonymized because they are blurred. 

After training all the baseline and proposed methods including GAN, features (representation) of labeled video were extracted. Then, a supervised (MLP-based) human activity recognition method was trained on features and the average top-1 classification accuracy on test set was calculated by using 5-fold cross validation and presented in Table \ref{results}. Baseline methods include GAN \cite{good16} and  self-supervised GAN with only rotation as learning task (GAN+Rotation) \cite{chen18} and proposed methods are self-supervised GAN with three different spatial transformations such as rotation, translation and shearing (GAN+Spatial), self-supervised GAN with only temporal transformation (shuffling) of video frames (GAN+Temporal) and finally self-supervised GAN with both spatial and temporal transformations (GAN-SpatioTemporal). 

The experimental results prove superiority of the proposed method (GAN-SpatioTemporal) over baseline GAN and GAN+Rotation for providing a useful representation of videos, specially for Ball-Drop dataset which is the focus of this paper. It is also interesting to see that in UCF101 dataset, GAN+Temporal outperforms GAN+Spatial and even GAN-SpatioTemporal

Next an ablation study is performed in order to investigate the effect of different spatial transformation used in proposed method on downstream classification accuracy. 
First, proposed method was trained using only one spatial transformation (rotation, translation or shearing). Then two transformation were used followed by all three. The top 1 classification accuracy of using features extracted from these methods applied on Ball-Drop dataset is shown in Table \ref{ablation}. The results shows that although rotation outperforms other transformation such as translation and shearing when used alone but combining different spatial transformation gains the best result.

\section{Conclusion and Future Works}
In this work, a novel method was proposed to augment GAN with a self-supervised task in order to improve its ability for generating useful representation of videos. The self-supervised task in this method consists of randomly picking a  transformation and applying it on video frames. Subsequently, the discriminator is encouraged to predict the correct transformation that was used. The experimental results proved that in overall, introduction of new transformations in different modalities enhances the capability of baseline GAN \cite{good16} and outperforms rotation only self-supervised GAN \cite{chen18} in providing a representation of videos useful for human activity recognition.

Next step would be using much deeper networks and applying this method on very large datasets, something that was beyond our computational budget at the moment. Another possible direction is to consider transformations as special case of policy in reinforcement learning \cite{RL}. Ability to find the policy that changed the state of the environment would be very useful in model-based reinforcement learning \cite{kais19}.

\begin{acks}
This  work  was  partially  supported  by  National  Science Foundation grants IIS 1565328 and IIP 1719031.
\end{acks}

\bibliographystyle{ACM-Reference-Format}
\bibliography{sample-base}


\end{document}